\documentclass{article}
\usepackage{ijcai24}

\usepackage{times}
\usepackage{soul}
\usepackage{url}
\usepackage{color}
\definecolor{blue}{RGB}{60,132,196}
\definecolor{red}{RGB}{207,78,56}
\definecolor{gray}{RGB}{146,146,161}
\definecolor{green4}{RGB}{46, 139, 87}
\usepackage[pagebackref,breaklinks,colorlinks,citecolor=green4]{hyperref}
\usepackage[utf8]{inputenc}
\usepackage[small]{caption}
\usepackage{graphicx}
\usepackage{amsmath}
\usepackage{amsthm}
\usepackage{booktabs}
\usepackage{algorithm}
\usepackage{algorithmic}
\usepackage[switch]{lineno}

\usepackage{graphicx}
\usepackage{amsmath}
\usepackage{amssymb}

\usepackage{booktabs}

\usepackage{multirow}
\DeclareMathOperator*{\argmax}{argmax}
\usepackage{dsfont}
\usepackage[mathscr]{eucal}

\usepackage{marvosym}
\usepackage{ifsym}
\usepackage{float}
\usepackage{subfigure}
\usepackage{colortbl}
\usepackage{xcolor}

\newcommand{\ie}{\textit{i.e.}}
\newcommand{\eg}{\textit{e.g.}}
\newcommand{\vs}{\textit{v.s.}}
\newcommand{\wrt}{\textit{w.r.t.}}
\newcommand{\State}{\STATE}
\definecolor{pink}{RGB}{115, 140, 166}
\definecolor{blue}{RGB}{60,132,196}
\definecolor{red}{RGB}{182,48,28}
\definecolor{gray}{RGB}{39, 88, 107}
\definecolor{green4}{RGB}{0, 152, 152}

\title{Evolutionary Generalized Zero-Shot Learning}

\author{
Dubing Chen$^1$
\and
Chenyi Jiang$^1$\and
Haofeng Zhang$^{1}$\thanks{Corresponding author}
\affiliations
$^1$School of Artificial Intelligence, Nanjing University of Science and Technology\\
\emails
\{db.chen, JiangChenyi, zhanghf\}@njust.edu.cn
}

\begin{document}

\maketitle

\begin{abstract}

Attribute-based Zero-Shot Learning (ZSL) has revolutionized the ability of models to recognize new classes not seen during training. However, with the advancement of large-scale models, the expectations have risen. Beyond merely achieving zero-shot generalization, there is a growing demand for universal models that can continually evolve in expert domains using unlabeled data. To address this, we introduce a scaled-down instantiation of this challenge: Evolutionary Generalized Zero-Shot Learning (EGZSL). This setting allows a low-performing zero-shot model to adapt to the test data stream and evolve online. We elaborate on three challenges of this special task, \ie, catastrophic forgetting, initial prediction bias, and evolutionary data class bias. Moreover, we propose targeted solutions for each challenge, resulting in a generic method capable of continuous evolution from a given initial IGZSL model. Experiments on three popular GZSL benchmark datasets demonstrate that our model can learn from the test data stream while other baselines fail. Codes are available at \url{https://github.com/cdb342/EGZSL}.

\end{abstract}

\section{Introduction}
\label{sec:intro}
In the era of large-scale models, it is critical that systems learn autonomously from data without human supervision, generalize to new concepts, and minimize data-induced biases \cite{radford2021learning,ferrara2023should,burns2023weak}. Traditional attribute-based zero-shot learning (ZSL) \cite{lampert2009learning,farhadi2009describing} has instantiated these challenges on a smaller scale by utilizing attributes as intermediaries that enable models to recognize novel categories. However, conventional ZSL paradigms primarily engage in training static models, which struggle to correct prediction biases from unseen concepts and adapt to varying dynamic demands. Consequently, we ponder how a model trained on limited data can dynamically, cost-effectively, and efficiently self-evolve when exposed to data on novel concepts during deployment, making better decisions on unfamiliar concepts while maintaining its core capabilities.

In this paper, we build on the foundational principles of attribute-based ZSL and introduce a new setting: Evolutionary Generalized Zero-Shot Learning (EGZSL). We envisage a ZSL model that, after its initial training, can continually learn from a stream of unlabeled data. This model is designed to autonomously identify and adapt to unseen concepts, thereby evolving in conjunction with its underlying knowledge.

\begin{figure*}[t]
	\centering
		\includegraphics[width=0.83\textwidth,trim={0cm 0 0 0},clip]{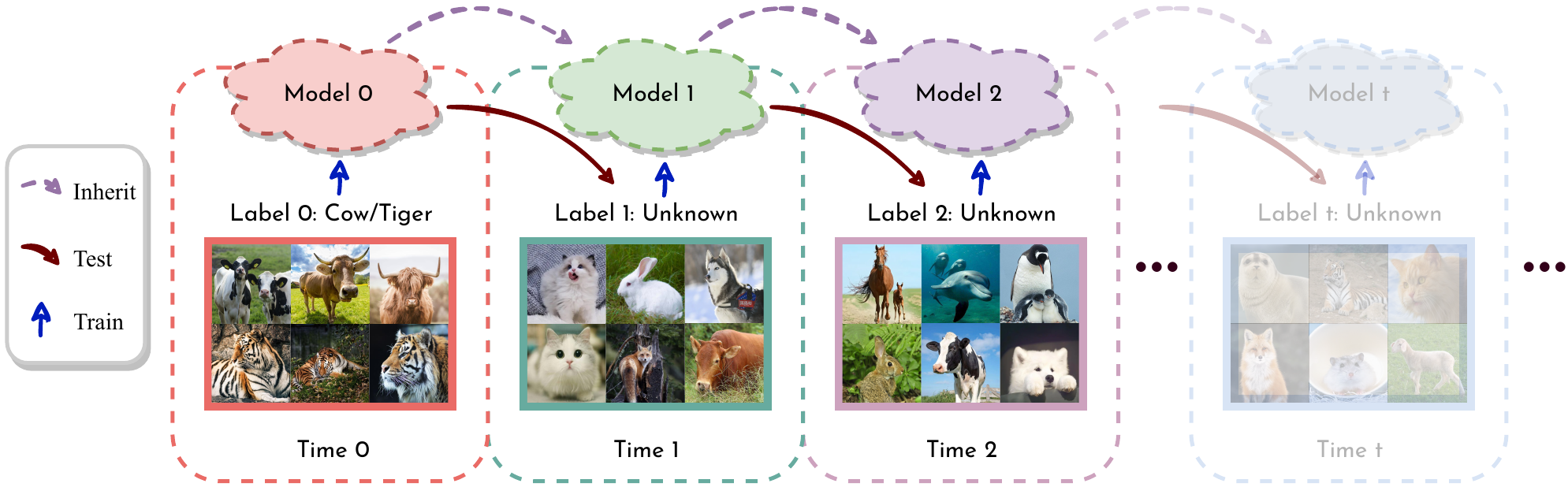}
	\caption{Illustration of the proposed EGZSL setting, featuring training with labeled seen class samples at time 0, followed by iterative predictive re-training on randomly divided data from the mixture of seen and unseen test sets in subsequent time steps (ratio of classes in a small batch is undefined). \textbf{Train:} train the current model at each time step with only the data indicated by the arrows. \textbf{Test:} predict the current data with the model obtained in the last time step. \textbf{Inherit:} train the current model based on the model of the last time step.}
    \label{fig:intro}
	\vspace{-0.5ex}
\end{figure*}

Distinct from existing ZSL settings (\ie, inductive ZSL (IGZSL) \cite{chao2016empirical,xian2017zero} and transductive ZSL (TGZSL) \cite{kodirov2015unsupervised,paul2019semantically,wan2019transductive,narayan2020latent}), EGZSL allows for unsupervised online enhancement during deployment, enabling the model to perpetually evolve through exposure to an unlabeled test data stream. This makes EGZSL \textit{(i) mitigate the domain shift problem \cite{fu2014transductive} by exposing the model to previously unseen class samples;} and \textit{(ii) suitable for real-world deployment.} Fig. \ref{fig:intro} briefly depicts the training and testing process of the proposed setting. At time 0, a base model is trained with the same settings as in IGZSL. At each subsequent time $t$, the model from time $t-1$ is first tested on the current batch, followed by unsupervised evolution. Unlike continual ZSL \cite{AGEM,gautam2020generalized}, we do not assume a fixed ratio of seen-unseen classes in each batch. Instead, data in each batch is randomly sampled from a mixture of both seen and unseen test sets. The model can only access the current data stream without having access to the base training data or the test data at other time stamps.

EGZSL meets three main challenges. First, the model is prone to catastrophic forgetting \cite{mccloskey1989catastrophic,french1999catastrophic} when training on streaming data. Second, due to the lack of unseen class samples in the base training phase, the prediction bias of the model is easily and consistently amplified when trained on the unlabeled data stream. Third, the model is vulnerable to potential data class imbalance problems. We then propose specific approaches to address these challenges. The overall framework is based on pseudo-label learning \cite{lee2013pseudo,xie2020unsupervised}, which is a common self-training approach for limited-supervised learning. We avoid forgetting by maintaining a global model, updated as the exponential moving average of the per-stage model. Historical information of the current model can be preserved by distilling it from the global model. We specify updateable class-related parameters for the class imbalance problem based on the data classes that occurred each time step. This avoids the error accumulation that causes predictions to deviate from certain classes. Moreover, we prevent confirmation bias by filtering noise labels. To avoid the effect of the initial prediction bias problem, we set a class-independent filtering threshold for each class. Finally, we propose evaluation criteria for this novel task, along with four baselines. The effectiveness of our proposed approach is demonstrated on three public ZSL benchmarks, on which the performance of our approach surpasses the base IGZSL method, while other baselines fail. Our contributions are summarized as follows:
\begin{itemize}
\item We establish a practical yet challenging evolutionary generalized zero-shot learning task, that is better suited for real-world applications than existing ZSL settings.
\item We analyze the main challenges of EGZSL and propose a targeted approach to address each of them.
\item We determine the evaluation criteria for EGZSL and conduct extensive experiments on three public ZSL datasets. The proposed method consistently improves over baselines. The effectiveness of our method is demonstrated by a series of explanatory experiments.
\end{itemize}

\section{Related Work}

\noindent \textbf{Zero-Shot Learning (ZSL)} \cite{lampert2009learning,lampert2013attribute,xian2017zero} aims at recognizing unseen classes when given only seen class samples. Early approaches \cite{akata2013label,elhoseiny2013write,frome2013devise}. typically embedded images and semantic descriptors (\eg, attributes, word vectors) to the same space, then conduct a nearest neighbor search. However, these methods were sensitive to the domain shift problem \cite{fu2014transductive}. They performed especially poorly in the \textbf{Generalized Zero-Shot Learning (GZSL)} setting \cite{chao2016empirical,xian2017zero}, which requires classifying both seen and unseen classes in the test phase. In the follow-up research, \cite{xian2018feature,xian2019f,shen2020invertible,han2021contrastive,chen2023deconstructed} employed conditional generative models \cite{kingma2013auto,arjovsky2017wasserstein,dinh2014nice} to generate pseudo-unseen class samples, thereby transferring GZSL into a supervised task. \cite{atzmon2019adaptive,chou2020adaptive} distinguished seen or unseen classes with out-of-distribution detectors \cite{fang2022out}, then classified in the corresponding subset of classes. \cite{xu2020attribute,jiang2024estimation} emphasized learning a deep embedding model.

\noindent \textbf{Transductive Zero-Shot Learning (TZSL)} \cite{kodirov2015unsupervised,wan2019transductive,narayan2020latent} assumed the unlabeled unseen test data is available during training. As a distinction, the earlier setting is called inductive ZSL. Existing methods \cite{fu2014transductive,bo2021hardness} typically relied on pseudo-labeling strategies.~\cite{xian2019f,narayan2020latent} also employed generative models. TZSL is a variant of semi-supervised learning \cite{grandvalet2004semi} on ZSL. It mitigates the domain shift problem and yields better recognition performance. However, for ZSL, the accessibility of unseen class samples in training is a too strong hypothesis, leading to limited application scenarios.

\noindent \textbf{Continual Zero-Shot Learning (CZSL)} \cite{AGEM,gautam2020generalized,skorokhodov2021class,yi2021domain} extended traditional ZSL into a class-incremental paradigm \cite{rebuffi2017icarl}. A-GEM \cite{AGEM} marked the inception of lifelong learning within the ZSL framework, exploring the efficacy of continuous learning methods and introducing a pragmatic evaluation protocol where each example is encountered only once. \cite{wei2020lifelong} refined this setup, proposing Lifelong Zero-Shot Learning, which sequentially learns from all seen classes across multiple datasets and evaluates on unseen data with the learned model. \cite{skorokhodov2021class} extended it to unlimited label searching space and let the model recognize unseen classes sequentially. Many subsequent methods in CZSL  have built upon this framework, with a focus on enhancing performance \cite{ghosh2021dynamic} or adapting to diverse applications. \cite{yi2021domain} extended the paradigm to various domains such as painting and sketching, introducing domain-aware continual zero-shot learning. In contrast, our EGZSL begins with a model trained on seen class samples, which are no longer accessible during the evolutionary process. We target to recognize forthcoming data streams comprising both seen and unseen classes, iteratively refining the model's recognition capability over time.

\noindent \textbf{Test Time Adaptation (TTA)} \cite{sun2020test,liu2021ttt++,wang2020tent,wang2022continual} enables the model to better adapt the test domain by constructing self-supervised learning (SSL) tasks on test data. This concept has been further developed into an online framework for continual refinement. \cite{sun2020test} employed a rotation prediction task to update the model in test time, which also served as an auxiliary task in training. \cite{liu2021ttt++} assessed TTA performance across various distribution shifts, and proposed to adopt contrastive learning as the SSL task. \cite{wang2020tent} removed the auxiliary task during training and employed the minimum entropy strategy for optimization during testing. Existing TTA research has focused chiefly on the distribution shift task, \ie, domain adaptation \cite{wang2020tent,wang2022continual}. We adapt TTA's strategy of unsupervised self-training during the testing phase to extend traditional ZSL, providing a more realistic setting than TGZSL and mitigating the domain shift problem in IGZSL. Due to the extreme class imbalance problem in the base training phase, the proposed EGZSL faces specific challenges.

\section{EGZSL: Settings and Challenges}
\begin{figure}[t]
	\flushleft 
		\includegraphics[width=0.47\textwidth]{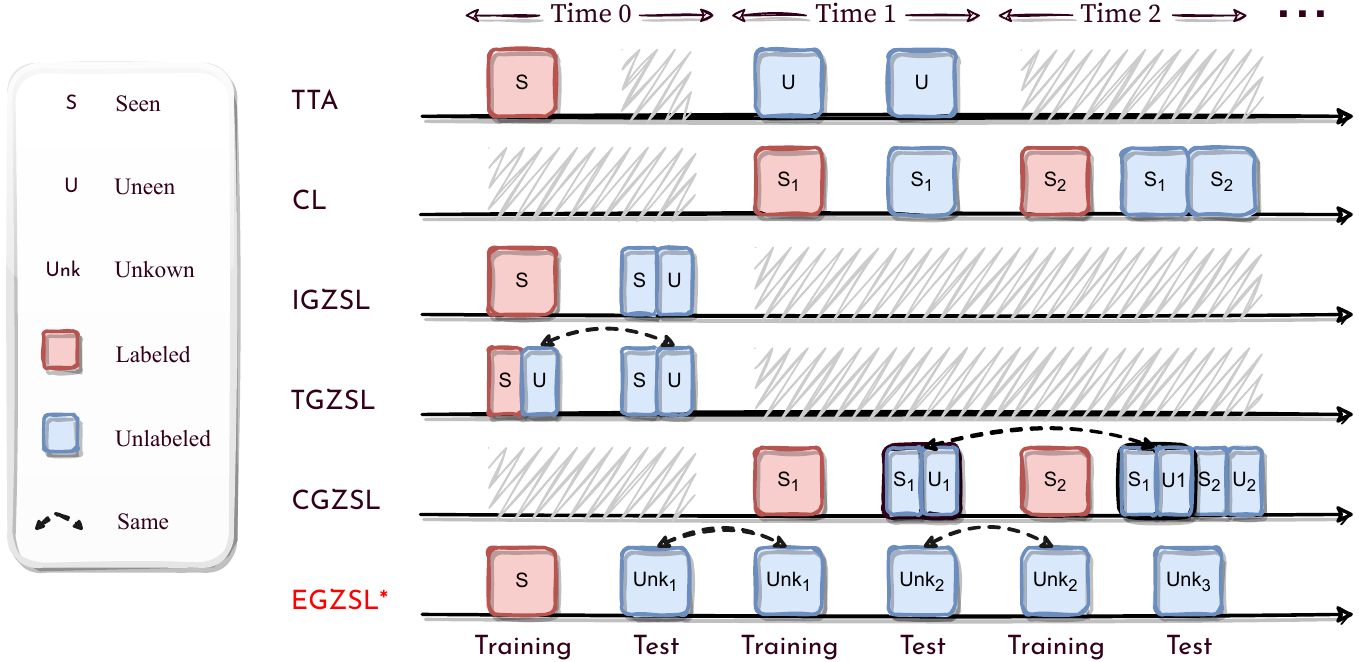}
	\caption{Comparasion of \textbf{EGZSL} with other similar settings in chronological progression. \textbf{TTA:} Test-Time Adaptation; \textbf{CL:} Continual Learning; \textbf{IGZSL:} Inductive Generalized Zero-Shot Learning; \textbf{TGZSL:} Transductive Generalized Zero-Shot Learning; \textbf{CGZSL:} Continual Generalized Zero-Shot Learning. In TTA, seen represents the source domain, and unseen is the target domain. In other settings, the labeled classes that appear in the training set are denoted as seen, and vice versa are unseen. An unknown class means that it can be any class (in seen or unseen classes).}
    \label{fig:comp}
	\vspace{-0.5ex}
\end{figure}

In this section, we formulate the EGZSL setting, analyze its key challenges, and compare it to other limited-supervision or incremental learning tasks. Fig. \ref{fig:comp} illustrates the differences between EGZSL and other similar settings.

\subsection{Problem Formulation}
EGZSL aims to evolve continually from a data stream. Let $\mathcal{Y}^{s}$ and $\mathcal{Y}^{u}$ denote two disjoint class label sets ($\mathcal{Y}=\mathcal{Y}^s\cup\mathcal{Y}^u$). $\mathcal{X} \subseteq \mathbb{R}^{d_\mathbf{x}}$ and $\mathcal{A} \subseteq \mathbb{R}^{d_\mathbf{a}}$ are feature space and attribute space, respectively, and $d_x$ and $d_a$ are dimensions of these two spaces. ZSL conventionally entails acquiring the associative relationship between visual features and semantic attributes to facilitate the transfer of knowledge to previously unseen classes. the goal of traditional GZSL is to learn such a classifier, \ie, $f_{gzsl}:\mathcal{X}\rightarrow\mathcal{Y}^{s}\cup\mathcal{Y}^u$ given the training set $\mathcal{D}^{tr}=\{ \mathbf{x},y |\mathbf{x}\in \mathcal{X},y\in \mathcal{Y}^{s}\}$ and the global semantic set $\mathcal{A}$.

In the initial phase (time 0), EGZSL endeavors to learn a foundational model \(f_{0}\) using a base set \(\mathcal{D}^{b}=\{ \mathbf{x}_i,y_i,{\mathbf{a}}_{y_i} |\mathbf{x}_i\in \mathcal{X},y_i\in \mathcal{Y}^{s},{\mathbf{a}}_{y_i}\in \mathcal{A}\}_{i=1}^{N_b}\), where \(N_b\) represents the volume of data in the base set. The base model \(f_{0}\) inherently possesses the capability to distinguish between both seen and unseen classes, denoted as \(f_{0}:\mathcal{X}\rightarrow\mathcal{Y}\). Subsequently, in time \(1, ..., t, ..., T\), the base model undergoes testing and evolution on the test data streams \(\mathcal{D}^{te}_{1}, ..., \mathcal{D}^{te}_{t}, ..., \mathcal{D}^{te}_{T}\), resulting in \(f_{1}, ..., f_{t}, ..., f_{T}\), where \(\mathcal{D}^{te}_{t}=\{ \mathbf{x}_j,y_j|\mathbf{x}_j\in \mathcal{X},y_j\in \mathcal{Y}\}_{j=1}^{N_t}\). Here, \(N_t\) denotes the data volume at time \(t\). Notably, \(\mathcal{D}^{te}_{t}\) is tested with \(f_{t-1}\), and \(f_{t-1}\) is subsequently retrained with the unlabeled data in \(\mathcal{D}^{te}_{t}\). The objective for \(f_t\) is to exhibit improved performance compared to \(f_{t-1}\) in classifying data with labels in \(\mathcal{Y}\). Training in the labeled base set and the unlabeled sequential test set is referred to as base learning and evolutionary learning, respectively. Ultimately, EGZSL performance is assessed based on the test results across all test subsets.

\subsection{Challenges Analysis}
\label{sec:ca}
\noindent \textbf{Catastrophic forgetting problem.} When training on the one-time given data, it is able to repeatedly utilize the data to achieve the global optimum. On the contrary, there is a contradiction between falling into a local optimum and underutilization of data when dealing with incremental data streams. Since only a small amount of data can be accessed at a time, overfitting on this batch of data will lead to catastrophic forgetting \cite{mccloskey1989catastrophic,french1999catastrophic} the previously learned knowledge. Conversely, the batch of data cannot be fully utilized, resulting in low training efficiency. Finding a solution that balances efficient data utilization with preventing forgetting is imperative.

\noindent \textbf{Initial bias problem.} Unlabeled training on evolving data is heavily influenced by the accuracy of pseudo-labels predicted by the base model. However, when the base model is trained on imbalanced data classes, its prediction will be biased towards specific classes. This problem exists in EGZSL since the base set lacks unseen class samples. The prediction imbalance problem will cause error accumulation when continuing training with unbalanced pseudo-labels.

\noindent \textbf{Sensitivity to data class bias.} Since the EGZSL setting assumes arbitrary class distributions for the evolutionary learning phase, the model is at risk of being exposed to class-biased batch data. Training on biased data can lead to bias in the model predictions for pseudo-labeling the next phase, which in turn increases the model bias, ultimately causing a progressive impact on the sequential data.
\subsection{EGZSL \vs Similar Settings}
As shown in Fig. \ref{fig:comp}, we compare EGZSL to existing settings. In detail, \textbf{IGZSL} depicts a static model that is trained once on seen classes, which does not adapt over time. \textbf{TGZSL} is similar to IGZSL but integrates (unlabeled) unseen class data during training. \textbf{TTA} adapts at test time, but the model trains on a fixed set of classes and it does not continue to learn or adapt beyond time. \textbf{CL} progressively trains the model on different subsets of classes, testing it on both new and previously learned classes, challenging the model to remember old knowledge.
\textbf{CGZSL} extends CL by including unseen classes in tests, pushing the model to adapt to new information constantly. Moreover, the proposed \textbf{EGZSL} raises the stakes, requiring the model to identify new classes, learn from ongoing unlabeled data flows, make sense of data without labels, and work without setting class limits at each time step. To put it in another perspective, EGZSL can be regarded as a strict version of TGZSL. TGZSL assumes that all labeled seen class data and unlabeled unseen class test data are given one-off, along with a fixed test set and known seen-unseen class splitting in the test set. In experiments, we consider IGZSL and TGZSL as upper and lower bounds for EGZSL to evaluate its performance since there is no existing method for EGZSL.

\section{Method}
EGZSL extends the IGZSL setting in test time. Any existing IGZSL models can be employed in our setting without retraining on the base data. Since base learning has been well studied, we focus mainly on the evolutionary learning phase. This section describes our method and explains each component that promotes evolutionary learning.

At each time step $t$, we first predict the pseudo-label of current data. To prevent catastrophic forgetting, we maintain a momentum model to preserve global data information and distill it to the current model. In addition, we select learnable class parameters to prevent class imbalance learning and filter unreliable data to avoid error accumulation. A specific training process is described in Algorithm \ref{algo}.
\subsection{Training with Pseudo Labels}
\label{sec:psudo_label}
Suppose a base model has been trained on the base set. We employ a pseudo-labeling strategy \cite{lee2013pseudo,xie2020unsupervised} to enable continual improvement from the unlabeled data, which is a typical technique in semi-supervised learning \cite{lee2013pseudo} and domain adaptation \cite{wang2020tent}. In time $t$, the pseudo label $\hat{y}_\mathbf{x}$ of a datum $\mathbf{x}$ is predicted with the highest compatibility with the model of the immediately preceding stage:
\begin{equation}
	\begin{split}
\hat{y}_\mathbf{x}=f_{t-1}(\mathbf{x})=\argmax_y F_{t-1}(\mathbf{x},y; \mathbf{W}_{t-1}).
	\end{split}
      \label{eq:pl}
\end{equation}
Here, $F_{t-1}$ measures the compatibility score between $\mathbf{x}$ and any class $y$, with $\mathbf{W}_{t-1}$ denoting its parameter. Based on the pseudo labels, we employ cross-entropy in label space $\mathcal{Y}$ to further optimize $\mathbf{W}_{t-1}$, \ie,

\begin{equation}
	\begin{split}
 \ell_{ce}(\mathbf{x})&=-\log p_{t-1}(\mathbf{x},\hat{y}_\mathbf{x};\mathcal{Y}),\\
      p_{t-1}(\mathbf{x},y;\mathcal{Y})=&\frac{\exp(F_{t-1}(\mathbf{x},y; \mathbf{W}_{t-1}))}{\sum_{c \in\mathcal{Y}}\exp(F_{t-1}(\mathbf{x},c; \mathbf{W}_{t-1}))}.
	\end{split}
      \label{eq:ce}
\end{equation}
\subsection{Maintenance of Global Information}
\label{sec:global_info}
A challenge of EGZSL is the unavailability of all evolutionary data at one time. Directly updating the model with gradient descent at a certain time step can lead to catastrophic forgetting \cite{mccloskey1989catastrophic,french1999catastrophic}, which is a typical difficulty in sequential learning. We resort to the momentum updated model as the surrogate of historical information, similar to MoCo \cite{he2020momentum,chen2020improved}. Formally, the parameters of the momentum model undergo updates as exponential moving averages (EMA) based on the parameters of the model from the previous time step. Denoting $F_{ema}$ as the momentum model with parameters $\mathbf{W}_{ema}$, at each time step $t$, $\mathbf{W}_{ema}$ is updated as
\begin{equation}
	\begin{split}
    \mathbf{W}_{ema}=m_1\cdot\mathbf{W}_{ema}+(1-m_1)\cdot
\mathbf{W}_t,
	\end{split}
      \label{eq:ema}
\end{equation}
where $m_1 \in [0, 1)$ is a smoothing factor. We update $\mathbf{W}_{ema}$ with the gradient detached. $\mathbf{W}_{ema}$ is considered to retain the global information from past time steps, exhibiting smoother changes than $\mathbf{W}_t$. We distill \cite{hinton2015distilling} this information into the current model with Kullback-Leibler (KL) divergence:
\vspace{-2ex}

\begin{equation}
	\begin{split}
   \ell_{kl}(\mathbf{x})=\sum_{y \in\mathcal{Y}}p_t(\mathbf{x},y;\mathcal{Y})\log \frac{p_t(\mathbf{x},y;\mathcal{Y})}{p_{ema}(\mathbf{x},y;\mathcal{Y})}.
	\end{split}
      \label{eq:kl}
\end{equation}

Here $p_t(\mathbf{x},y;\mathcal{Y})$ and $p_{ema}(\mathbf{x},y;\mathcal{Y})$ denote the probability distribution over the variable $y$. Note that at each time step $t$, $\mathbf{W}_{ema}$ is updated after $\mathbf{W}_{t}$. With the distillation loss, the model can learn from the current data stream while avoiding catastrophic forgetting of previous information. This creates conditions for balancing the data utilization efficiency.

\subsection{Class Selection for Stable Training}
\label{sec:class_sel}
As discussed in Sec. \ref{sec:ca}, the potential imbalance of data classes in the evolutionary stage can lead to unbalanced predictions by the model, resulting in error accumulation. This problem becomes more pronounced when the number of samples available at each time step is small. It is prone to missing samples in certain classes, and cross-entropy based on pseudo-hard labels may produce sharper constraints that cause model predictions to abruptly deviate from these classes. To address this, we propose selecting specific class parameters to update at each time step. Consider typical GZSL classifiers are implemented with a linear model, \ie, $\mathbf{W}$ is a matrix with $|\mathcal{Y}|$ rows and $d_\mathbf{x}$ columns:
\begin{equation}
	\begin{split}
    F(\mathbf{x},y;\mathbf{W}):=\mathbf{W}_y\cdot \mathbf{x}.
	\end{split}
      \label{eq:weight}
\end{equation}
To ensure smoother updates of the weights for each class, at time step $t$, we choose to update only the classes present in the pseudo labels, \ie,
\begin{equation}
	\begin{split}
\mathcal{Y}_t^{sel}=\mathrm{unique}(\{\hat{y}_\mathbf{x}\}_{\mathbf{x} \in \mathcal{D}_t^{te}}).
	\end{split}
      \label{eq:select_class}
\end{equation}
Here $\mathrm{unique}(\cdot)$ denotes the function that returns the unique elements of the label set, which can be achieved by directly calling the \textit{PyTorch} function. Consequently, the cross-entropy loss in Eq. (\ref{eq:ce}) is substituted with
\begin{equation}
	\begin{split}
 \ell_{ce}^{sel}(\mathbf{x})&=-\log p_{t-1}(\mathbf{x},\hat{y}_\mathbf{x};\mathcal{Y}_t^{sel}).
	\end{split}
      \label{eq:sel_ce}
\end{equation}
Note that Eq. (\ref{eq:kl}) is still computed with the full label set $\mathcal{Y}$.
\subsection{Data Selection for Effective Training}
\label{sec:data_sel}
Since the unlabeled data is trained using the pseudo-labeling strategy, noisy pseudo-labels will introduce confirmation bias. Hence, at each time step, we employ the model from the previous stage to select samples with low uncertainty. The uncertainty reflects the confidence level of the model prediction, typically measured by entropy or max softmax prediction \cite{mukhoti2021deep}. We use the latter to select samples with more reliable pseudo labels. Intuitively, we can establish a constant threshold to filter the samples where softmax prediction fails to surpass this value, \ie,
\begin{equation}
	\begin{split}
   \texttt{M}(\mathbf{x})=\mathds{1}(\max_yp_{t-1}(\mathbf{x},y;\mathcal{Y})>\tau),
	\end{split}
      \label{eq:select}
\end{equation}
where $\mathds{1}$ denotes the indicator function, and $\tau \in (0,1]$ is the predefined threshold. $\texttt{M}(\cdot)$ enables the selection of the samples with high confidence. However, when softmax prediction values are unbalanced across classes, employing a fixed threshold results in unbalanced data selection. For instance, the base model is trained only on the seen classes, exhibiting higher confidence in samples from seen classes and lower confidence in those from unseen classes. Using the fixed threshold approach could lead to filtering out too many unseen class samples. This selection imbalance in sequential data learning can give rise to the Matthew Effect.

We adopt an adaptive threshold for each class to address this problem. Recognizing that the imbalance in selection arises from variations in softmax prediction distributions across classes, we leverage class statistics to establish class-independent thresholds. Specifically, we incorporate a curriculum learning strategy \cite{bengio2009curriculum} to consider the learning progress of each class. Given the limited number of samples available at each time step, we aggregate statistics across all historical time steps to compute class confidence statistics.  These statistics are momentum updated as follows:
\begin{equation}
	\begin{split}
   \delta_{ema}(y)&=m_2\cdot \delta_{ema}(y)+(1-m_2)\cdot \delta_{t-1}(y),\\
  \delta_{t-1}(y)&=\frac{1}{N_t^y}\sum_{\mathbf{x} \in \mathcal{D}^{te}_{t}}\mathds{1}(y=\hat{y}_\mathbf{x})  p_{t-1}(\mathbf{x},\hat{y}_\mathbf{x};\mathcal{Y}),
	\end{split}
      \label{eq:thre}
\end{equation}
where $m_2 \in [0, 1)$ denotes a momentum coefficient, and $N_t^y=\sum_{\mathbf{x} \in \mathcal{D}^{te}_{t}}\mathds{1}(y=\hat{y}_\mathbf{x})$ ($\hat{y}_\mathbf{x}$ is defined in Eq.~(\ref{eq:pl})). $\delta_{t-1}(y)$ represents the averaged softmax prediction for class $y$, predicted by the immediately preceding stage model. $\delta_{ema}$ serves as the surrogate of the class learning status and is updated before data selection at time $t$.  It is then utilized to adjust the fixed threshold $\tau$. The scaled data mask is
\begin{equation}
	\begin{split}
    \texttt{M}_{scl}(\mathbf{x})=\mathds{1}((\mathbf{x},f_{t-1}(\mathbf{x}))>\delta_{t-1}(f_{t-1}(\mathbf{x}))\cdot\tau).
	\end{split}
      \label{eq:thre_scale}
\end{equation}
$\texttt{M}_{scl}(\cdot)$ is subsequently employed as a weighting factor for the loss of each datum to facilitate data selection, \ie,
\begin{equation}
 \resizebox{0.89\linewidth}{!}{$
	\begin{aligned}
   \mathcal{L}_{ce}^{sel}=\mathbb{E}_{\mathbf{x} \in \mathcal{D}^{te}_{t}}\mathtt{M}_{scl}(\mathbf{x})\cdot\ell_{ce}^{sel}(\mathbf{x}), \mathcal{L}_{kl}=\mathbb{E}_{\mathbf{x} \in \mathcal{D}^{te}_{t}}\mathtt{M}_{scl}(\mathbf{x})\cdot\ell_{kl}(\mathbf{x}).
	\end{aligned}
             $}
      \label{eq:scl}
\end{equation}
Notably, the class and data selection processes only incur negligible extra computation.
\begin{algorithm}[t]
\caption{The Proposed EGZSL Method}
\label{algo}
    	\textbf{Input:}~
Subset of data~$\{\mathbf{x}_i\}_{i=1}^{N_t}$; Model~$f_{t-1}$ with parameters $\mathbf{W}_{t-1}$; Momentum model $f_{ema}$ with parameters $\mathbf{W}_{ema}$; Class momentum confidence $\delta_{ema}$; Hyper-parameters $m_1, m_2, \tau, \lambda$.
                    \begin{algorithmic}[1]
            \STATE Predict pseudo-label: $\hat{y}_\mathbf{x}=f_{t-1}(\mathbf{x})$.
            \State Select training classes by Eq. (\ref{eq:select_class}).
            \State Update class momentum confidence $\delta_{ema}$ by Eq. (\ref{eq:thre}).
            \State Calculate data mask by Eq. (\ref{eq:thre_scale}).
            \State Update~$\mathbf{W}_{t-1}$ by cross-entropy loss and KL divergence loss in~Eq.~(\ref{eq:all}).
            \State Update weights $\mathbf{W}_{ema}$ of  momentum model by Eq. (\ref{eq:ema}).
            		\end{algorithmic}
\textbf{Output:} Prediction $\{f_{t-1}(\mathbf{x})|\mathbf{x}\in \mathcal{D}_t^{te}\}$; Updated model $f_{t}$; Updated momentum model $f_{ema}$; Updated class momentum confidence $\delta_{ema}$.
 \end{algorithm}
\subsection{Overall Objectives}
Overall, the total objective loss function at each time $t$ is
\begin{equation}
	\begin{aligned}
   \mathcal{L}_{all}=\mathcal{L}_{ce}^{sel}+\lambda\mathcal{L}_{kl},
	\end{aligned}
      \label{eq:all}
\end{equation}
where $\lambda$ is a hyper-parameter for balancing loss $\mathcal{L}_{ce}^{sel}$ and $\mathcal{L}_{kl}$. Algorithm \ref{algo} describes the concrete training process in one evolutionary learning step.
      \begin{table*}[t]
\newcommand{\tabincell}[2]{\begin{tabular}{@{}#1@{}}#2\end{tabular}}
	\centering
	\resizebox{\textwidth}{!}{
		\begin{tabular}{clccc|ccc|ccc}
        \toprule
			&\multirow{2}{*}{Method}                                               & \multicolumn{3}{c}{AWA2}                                                       & \multicolumn{3}{c}{CUB}                                                                                                 & \multicolumn{3}{c}{APY}                      \\
			&                        & \multicolumn{1}{|c}{$\mathit{A}^u$}                           &$\mathit{A}^s$                    & $\mathit{H}$                         & $\mathit{A}^u$                        &$\mathit{A}^s$                      & $\mathit{H}$                      & $\mathit{A}^u$                     & $\mathit{A}^s$                    & $\mathit{H}$                         \\
\midrule
 \multicolumn{1}{c|}{\multirow{3}{*}{$\mathcal{T}$}}& \multicolumn{1}{l|}{COND \cite{li2019rethinking}}          &80.2 &90.0 &84.8 &57.0 &68.7& 62.3 &51.8 &87.6 &65.1\\      
  \multicolumn{1}{c|}{}& \multicolumn{1}{l|}{TF-VAEGAN \cite{narayan2020latent}}          &87.3 &89.6 &88.4 &69.9 &72.1 &71.0   &-&-&-\\ 
 \multicolumn{1}{c|}{}& \multicolumn{1}{l|}{STHS \cite{bo2021hardness}}          &94.9& 92.3 &93.6 &77.4 &74.5& 75.9&-&-&-\\         
\midrule
\midrule                 
                           \multicolumn{1}{c|}{\multirow{5}{*}{$\mathcal{I}$}}&  \multicolumn{1}{l|}{COND \cite{li2019rethinking}}            &56.4 &81.4 &66.7 &47.4 &47.6 &47.5 &26.5&74.0&39.0\\
                \multicolumn{1}{c|}{}&	 \multicolumn{1}{l|}{ FREE \cite{chen2021free}}                  & 60.4          & 75.4          & 67.1          & 55.7          & 59.9          & 57.7                 & -             & -             & -             \\

\multicolumn{1}{c|}{} &\multicolumn{1}{l|}{Chou et al. \cite{chou2020adaptive}} &65.1 &78.9 &71.3  &41.4 &49.7 &45.2    &35.1 &65.5 &45.7\\
    \multicolumn{1}{c|}{} &\multicolumn{1}{l|}{GCM-CF \cite{yue2021counterfactual} }       &60.4 &75.1 &67.0  &61.0 &59.7 &60.3      &37.1 &56.8 &44.9\\
 \multicolumn{1}{c|}{}&	\multicolumn{1}{l|}{ZLA \cite{chen2022zero}}                  &65.4	&82.2	&72.8          &50.9	&58.4	&54.4              & 38.4	&60.3	&46.9           \\
\midrule
\midrule
   \multicolumn{1}{c|}{}   &\multicolumn{1}{l|}{COND+ERM@10}              & ${51.9}_{\pm0.3}$& ${75.5}_{\pm0.2}$       & ${61.5}_{\pm0.3}$      & ${41.1}_{\pm0.4}$ &${45.4}_{\pm0.5}$&${43.1}_{\pm0.3}$     &  ${26.3}_{\pm0.4}$    & ${45.8}_{\pm0.8}$       &${33.4}_{\pm0.5}$    \\ 
        \multicolumn{1}{c|}{}   &\multicolumn{1}{l|}{COND+ERM@100}              & ${51.2}_{\pm0.3}$& ${74.7}_{\pm0.4}$       & ${60.1}_{\pm0.1}$      & ${36.3}_{\pm0.3}$ &${39.6}_{\pm 0.7}$&${37.9}_{\pm0.4}$     &  ${25.4}_{\pm1.1}$    & ${43.0}_{\pm0.6}$       &${32.0}_{\pm0.8}$    \\ 
        \rowcolor{pink!30}\multicolumn{1}{c|}{\cellcolor{white!0}}&	\multicolumn{1}{l|}{COND+\textbf{ours@10}  }     &${57.1}_{\pm0.5}$ &${\bf82.1}_{\pm0.1}$ &${67.4}_{\pm0.3}$     & ${\bf45.2}_{\pm0.1}$&${54.6}_{\pm0.1}$  &${49.4}_{\pm0.1}$     &  ${32.1}_{\pm0.7}$   &  ${\bf60.1}_{\pm0.2}$&  ${41.9}_{\pm0.6}$ \\ 
\rowcolor{pink!30}\multicolumn{1}{c|}{\cellcolor{white!0}}&	\multicolumn{1}{l|}{COND+\textbf{ours@100}  }                 &${\bf59.2}_{\pm1.1}$&${80.7}_{\pm0.4}$ &${\bf68.3}_{\pm0.8}$     & ${45.0}_{\pm0.2}$&${\bf55.2}_{\pm0.3}$  &${\bf49.6}_{\pm0.1}$     &  ${\bf35.2}_{\pm0.2}$   &  ${58.0}_{\pm0.6}$&  ${\bf43.8}_{\pm0.2}$ \\ 
\cmidrule{2-11}
      \multicolumn{1}{c|}{} &\multicolumn{1}{l|}{ZLA+ERM@10}              & ${54.2}_{\pm6.5}$& ${60.4}_{\pm3.1}$       & ${56.9}_{\pm4.2}$      & ${45.2}_{\pm4.3}$ &${42.7}_{\pm4.0}$&${43.9}_{\pm4.0}$     &  ${8.6}_{\pm0.2}$    & ${0.4}_{\pm0.3}$       &${0.9}_{\pm0.4}$    \\ 
        \multicolumn{1}{c|}{}   &\multicolumn{1}{l|}{ZLA+ERM@100}              & ${55.0}_{\pm3.1}$& ${64.6}_{\pm2.5}$       & ${59.4}_{\pm2.7}$      & ${\bf52.0}_{\pm0.7}$ &${51.2}_{\pm 0.7}$&${51.6}_{\pm0.6}$     &  ${11.5}_{\pm1.4}$    & ${5.1}_{\pm0.9}$       &${6.8}_{\pm0.5}$    \\ 
       \rowcolor{pink!30}\multicolumn{1}{c|}{\cellcolor{white!0}}&	\multicolumn{1}{l|}{ZLA+\textbf{ours@10}  }     &${65.4}_{\pm0.6}$ &${\bf85.8}_{\pm0.5}$ &${74.2}_{\pm0.2}$     & ${51.0}_{\pm0.3}$&${\bf58.9}_{\pm0.3}$  &${\bf54.6}_{\pm0.1}$     &  ${39.1}_{\pm1.1}$   &  ${\bf60.1}_{\pm1.1}$&  ${47.3}_{\pm0.8}$ \\

  \rowcolor{pink!30}\multicolumn{1}{c|}{\multirow{-8}{*}{\cellcolor{white!0}$\mathcal{E}$}} &	\multicolumn{1}{l|}{ZLA+\textbf{ours@100}  }                 &${\bf73.3}_{\pm1.0}$ &${81.3}_{\pm0.8}$ &${\bf77.0}_{\pm0.2}$     & ${51.7}_{\pm0.6}$&${57.9}_{\pm0.3}$  &${54.6}_{\pm0.3}$     &  ${\bf40.0}_{\pm1.0}$   &  ${58.6}_{\pm0.7}$&  ${\bf47.5}_{\pm0.8}$ \\ 
  
            \bottomrule
            \end{tabular}
            }
                	\caption{Performance comparison between the proposed baselines and sota IGZSL and TGZSL methods. $\mathcal{T}$, $\mathcal{I}$, and $\mathcal{E}$ denote methods in the TGZSL, IGZSL, and EGZSL settings, respectively. @10 and @100 indicate the amount of data accessed in a single evolutionary time step. $\mathit{A}^u$ and $\mathit{A}^s$ represent per-class accuracy scores (\%) in seen and unseen test sets, and $\mathit{H}$ is their harmonic mean. The best results are bolded.}
     \label{tab:main}
	\vspace{-1.5ex}
\end{table*}
\section{Experiments}
In this section, we propose a protocol for evaluating EGZSL methods and compare the performance of our method to potential upper and lower boundaries. We also report on further experiments that shed light on the working mechanisms of our method by isolating the effects of individual components.

\subsection{Benchmark Protocol}

\noindent \textbf{Evaluation Procedure.}~As there is no established benchmark protocol for assessing EGZSL performance, we propose the following evaluation procedure: for a given ZSL dataset, the original training set serves as the base set, while the test set is partitioned into various batches in a fixed random order. Each method is initially trained on the base set, followed by prediction and online updating on the divided test data stream. Predictions for the current time step are made by the model from the previous step. We present two variations of the test data splitting: dividing the test set into batches of 10 or 100 samples. To compare with traditional GZSL methods, we employ the same metrics \cite{xian2017zero} for the EGZSL task, computed as the harmonic mean ($\mathit{H}$) of the average per-class top-1 accuracies in the seen ($\mathit{A}^s$) and unseen ($\mathit{A}^u$) classes. EGZSL performance is evaluated by aggregating predictions across all test batches. Each benchmark is repeated five times with different random data orders, and averages along with standard deviations of the results are reported.

\noindent \textbf{Datasets.} We evaluate EGZSL methods on three public ZSL benchmarks: 1) \textit{Animals with Attributes 2 (AWA2)} \cite{lampert2013attribute} contains 50 animal species and 85 attribute annotations, accounting for 37,322 samples. 2) \textit{Attribute Pascal and Yahoo (APY)} \cite{farhadi2009describing} includes 32 classes of 15,339 samples and 64 attributes. 3) \textit{Caltech-UCSD Birds-200-2011 (CUB)} \cite{wah2011caltech} consists of 11,788 samples of 200 bird species, annotated by 312 attributes. We split the data into seen and unseen classes according to the common GZSL benchmark procedure in \cite{xian2017zero}. We follow \cite{xian2017zero} to adapt the 2048-dimensional visual representation (instead of the original images) extracted from the pre-trained ResNet101 \cite{he2016deep}.

\subsection{Implementation Details} 
\noindent \textbf{Model.}
As the base learning phase setup remains unchanged from IGZSL, we simply borrow the off-the-shelf IGZSL model for conducting EGZSL experiments. Our EGZSL approach is developed using linear classifiers (Eq. (\ref{eq:weight})) that were trained by COND \cite{li2019rethinking} and ZLA \cite{chen2022zero}. The base models are acquired with their official codes. Please refer to the original papers for details on the base phase training procedure.

\noindent \textbf{Optimization.}
We employ the Adam optimizer \cite{kingma2014adam} with a learning rate of $5e$-$5$ for the main experiments. We set the (mini) batch size equal to the total number of data in each evolutionary stage. Each stage of data is optimized for one epoch only.

\begin{figure}[t]
	\centering   
        \subfigure [Ours]
	{
		\includegraphics[width=0.2\textwidth]{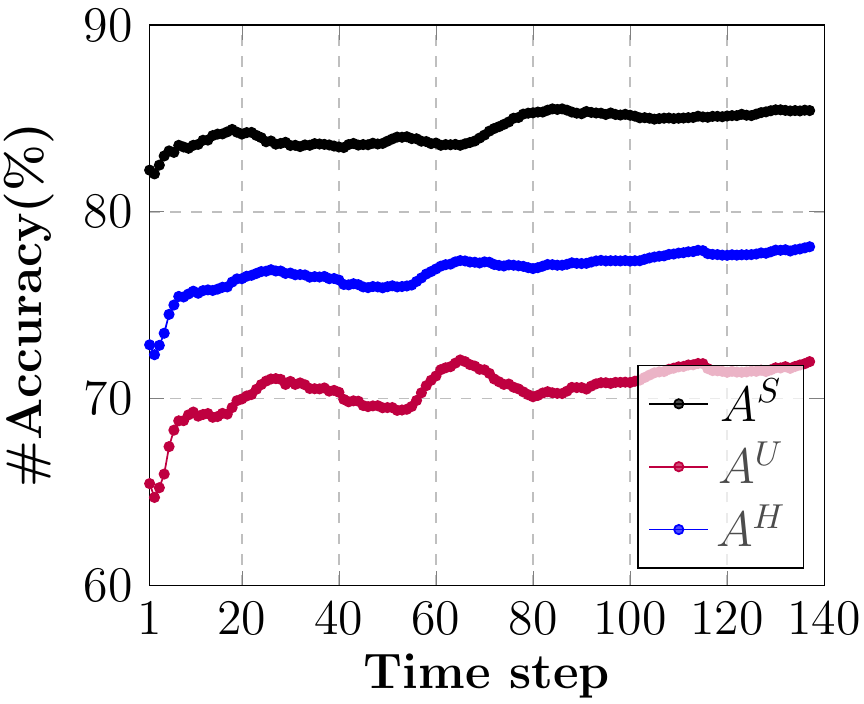}
	}
     \subfigure [ERM]
	{
		\includegraphics[width=0.2\textwidth]{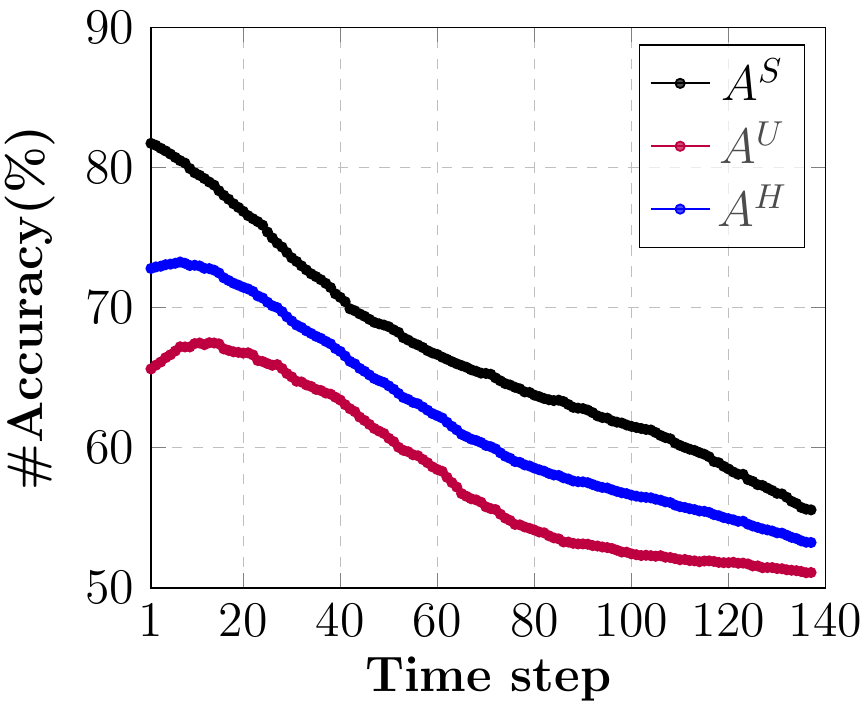}
	}
\vspace{-1.5ex}
	\caption{\textbf{(a), (b)} A comparison of evolution curves between our approach and ERM (on AWA2, with 100 samples per time step). Our method displays a rise in accuracy over time, while ERM experiences a decline in accuracy.}
	\label{fig:curve}
 \vspace{-1.5ex}
\end{figure}
\subsection{Main Results}
\label{sec:main_results}
\noindent \textbf{Baselines.} 
We establish our proposed method on COND \cite{li2019rethinking} and ZLA \cite{chen2022zero}, as shown in Tab. \ref{tab:main}. For comparison, we also report the results of a basic empirical risk minimization (ERM) approach with pseudo-labels. Additionally, we evaluate the performance of our approach against the IGZSL and TGZSL methods, which establish the potential high and low limits of the EGZSL performance.

\noindent \textbf{Results.} Tab.~\ref{tab:main} presents our main experimental results, according to which we have the following findings:

\textit{Our approach improves upon the IGZSL baseline, whereas the straightforward ERM approach fails.} For unsupervised data stream learning, relying on the basic method without addressing issues such as catastrophic forgetting, prediction bias, and data bias, results in unreliable gradients, especially when the initial model exhibits significant bias (as in the case of APY). In contrast, the superior results of our method demonstrate that it effectively handles these challenges.

\textit{The process of evolutionary learning provides greater benefits to coarse-grained datasets, such as AWA2 and APY, compared to fine-grained datasets like CUB.} This is attributed to the prevalence of visual bias \cite{fu2014transductive} resulting from missing unseen classes during base training, which is more pronounced in coarse-grained datasets. Consequently, mitigating this bias leads to a more substantial performance improvement. Additionally, the limited amount of evolutionary data per class (23 in CUB \vs 275 in AWA2) also contribute to the modest improvement observed in CUB. 

\textit{The performance improvement is slightly larger when more data is provided per time step.} Accessing more data at once lessens the probability of being overwhelmed by erroneous pseudo-label samples and presents a consistent gradient. In a real-world deployment, the batch size of a single evolutionary step can be selected according to specific requirements and resource usage.

\textit{Even though our approach outperforms the IGZSL baseline, there remains a significant gap between its results and those obtained by methods in the TGZSL setting.} This is primarily owing to three reasons: \textbf{first}, TGZSL allows for repeatedly training on all of the test data; \textbf{second}, TGZSL offers a more relaxed constraint by providing prior knowledge regarding whether the test data belong to seen or unseen classes; and \textbf{third}, greater attention has been paid to the TGZSL area, while there is still potential for further advancement in EGZSL. Regardless, the EGZSL setting is better suited for practical applications and holds greater potential for real-world deployment.

\subsection{Evolution Curves}
\label{sec:curve}
To evaluate model performance across different time steps, we plot its evolution curve as the evolutionary task progresses. Given that the test data differs among time steps, we compile the evolution curve using all test data. This is legitimate within the TGZSL setting and only applies to explanatory experiments. As shown in Fig. \ref{fig:curve}, our method demonstrates a consistent improvement in performance over time, whereas basic ERM leads to continuous forgetting of initial knowledge. This experiment is conducted based on ZLA.

\begin{table}[t]
	\centering
	\resizebox{0.49\textwidth}{!}{
\begin{tabular}{@{}llllllll@{}}
\toprule
&\multirow{2}{*}{Baseline}                      & \multicolumn{3}{c}{AWA2}                                                                                    & \multicolumn{3}{c}{APY}                                                                                     \\
                                      & & \multicolumn{1}{c}{$\mathit{A}^u$} & \multicolumn{1}{c}{$\mathit{A}^s$} & \multicolumn{1}{c}{$\mathit{H}$} & \multicolumn{1}{c}{$\mathit{A}^u$} & \multicolumn{1}{c}{$\mathit{A}^s$} & \multicolumn{1}{c}{$\mathit{H}$} \\ 
                                       \midrule
(i)&W/O Momentum Model              &  57.2                                        &    80.2                                      &   66.7                 
& 17.8                                         & 45.5                                         &25.6                       \\
(ii)&W/O Class Selection                    &   \textbf{66.4}                                       &  72.9                                        &      69.5                 &   \textbf{39.1}                                       &  51.3                                        &    44.4                   \\
(iii)&W/O Data Selection                     &     62.3                                     &  86.7                                   &  72.5                     &       38.7                                   &     57.3                                     &   46.2                    \\
(iv)&Adaptive$\rightarrow$fixed thre.                      &     57.4                                     &  \textbf{86.2}                                   &  68.9                     &       27.6                                   &     50.4                                     &   35.7                    \\
&\textbf{Full model}                             &$65.9$                                          & 85.5                                         & \textbf{74.4}                      & 38.6                                         & \textbf{61.5}                                        & \textbf{47.4}                      \\ 
\bottomrule
\end{tabular}
}
\caption{Ablation study results of the proposed method on AWA2 and APY datasets (with 10 samples per time step). }
\vspace{-1ex}
\label{tab:ablation}
\end{table}

\subsection{Ablation Analysis}
\label{sec:ablation}
We validated the effectiveness of our motivation and design through the following baselines, and the results are presented in Tab. \ref{tab:ablation}. These results are obtained by utilizing ZLA as the base model and maintaining a consistent random seed throughout the testing process.

\noindent \textit{(i) W/O Momentum Model.}~We first investigate the effect of the momentum model, which aids in preserving historical information. As shown in Tab.~\ref{tab:ablation}, omitting this component leads to performance degradation on both datasets. The decline in performance is particularly pronounced when the initial accuracy is low (as observed in APY). In the absence of historical information, noisy pseudo-labels dominate the training process. This demonstrates the importance of suppressing catastrophic forgetting in training with streaming data.

\noindent \textit{{(ii) W/O Class Selection.}}
We also evaluate the importance of the class selection module. As previously discussed, this module helps mitigate the adverse effects of potentially imbalanced data classes. There is a noticeable decrease in performance when ablating it.

\noindent \textit{{(iii) W/O Data Selection.}}
This baseline removes the operations defined by Eq. (\ref{eq:thre}), (\ref{eq:thre_scale}), and (\ref{eq:scl}). The removal of data selection implies that all samples are involved in the training. The decline in performance aligns with our expectation that filtering out low-confidence samples is beneficial.

\noindent \textit{{(iv) Adaptive$\rightarrow$fixed thre.}} To demonstrate the effectiveness of the adaptive threshold strategy in data selection, we conduct an experiment with a fixed threshold. This baseline is also described in Sec. \ref{sec:data_sel} that replaces Eq. (\ref{eq:thre_scale}) with Eq. (\ref{eq:thre}). We set $\tau$ to 0.8 for the best results of this baseline, but the performance is even worse than without data selection. This validates our analysis that a fixed threshold comes with the risk of data class imbalance.

\subsection{Hyperparameters} \label{sec:hyperparameters}

\begin{figure}[t]
	\centering
        \subfigure []
	{
		\includegraphics[width=0.2\textwidth]{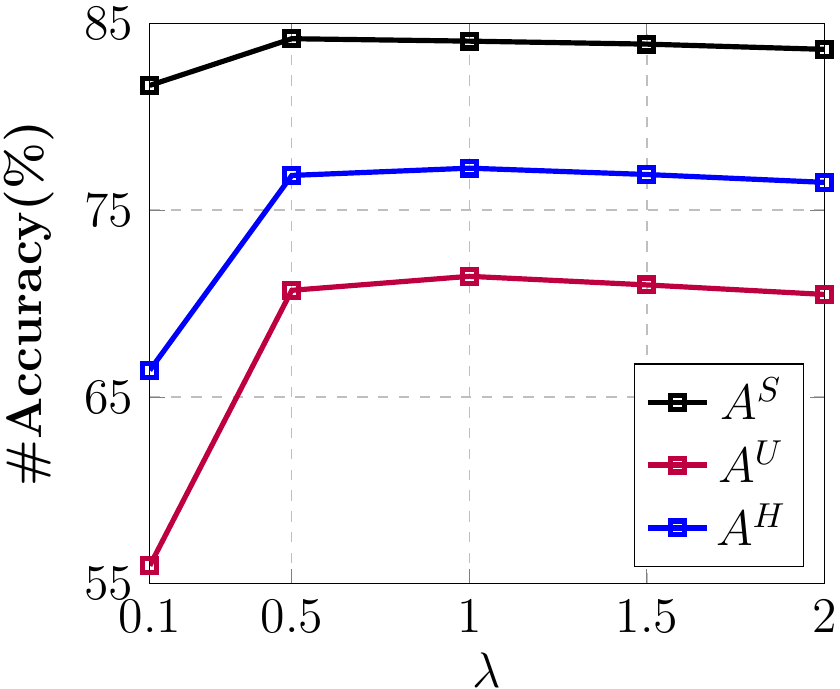}
	}
     \subfigure []
	{
		\includegraphics[width=0.2\textwidth]{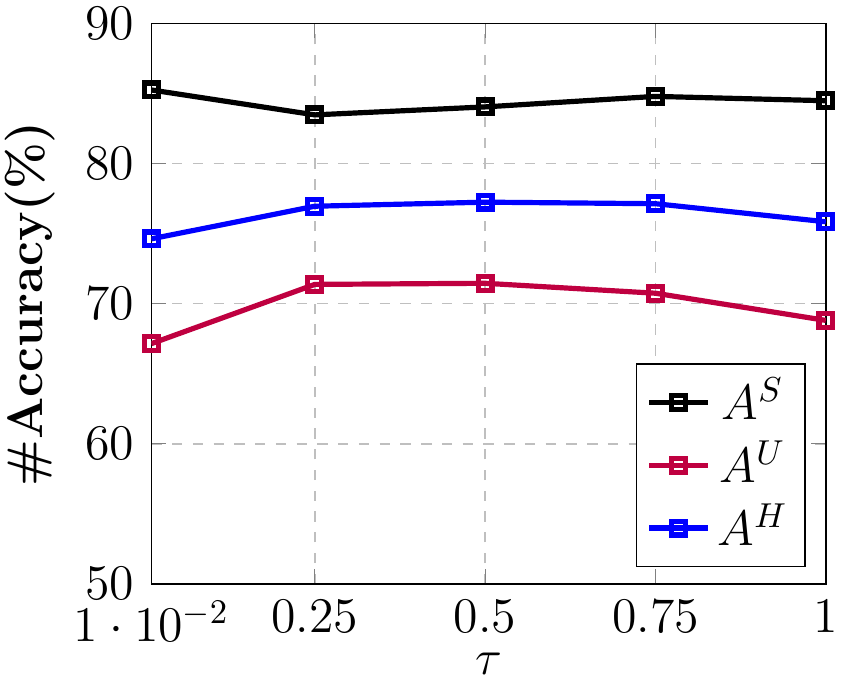}
	}\\
	\vspace{-2ex}
            \subfigure []
	{
		\includegraphics[width=0.2\textwidth]{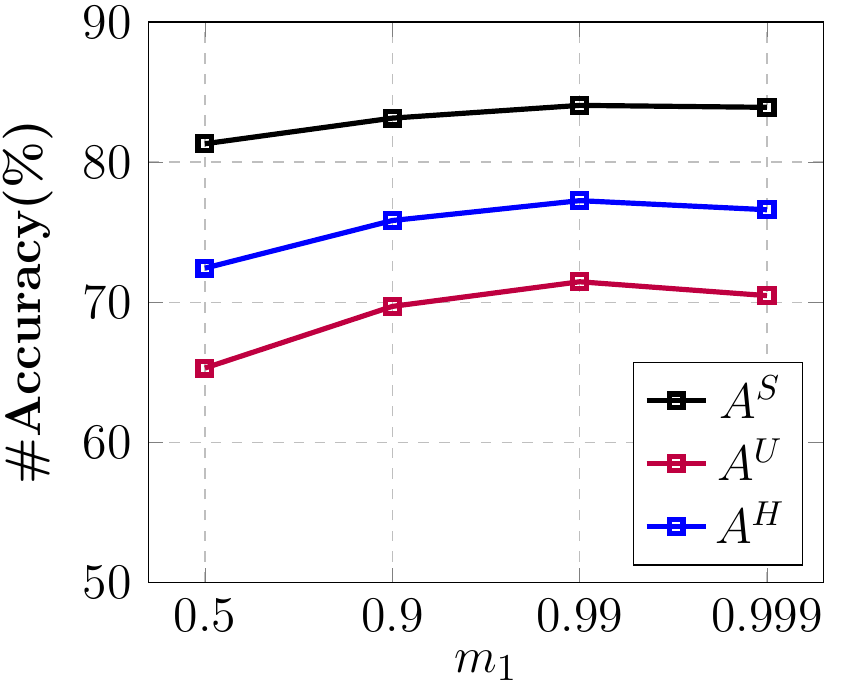}
	}
     \subfigure []
	{
		\includegraphics[width=0.2\textwidth]{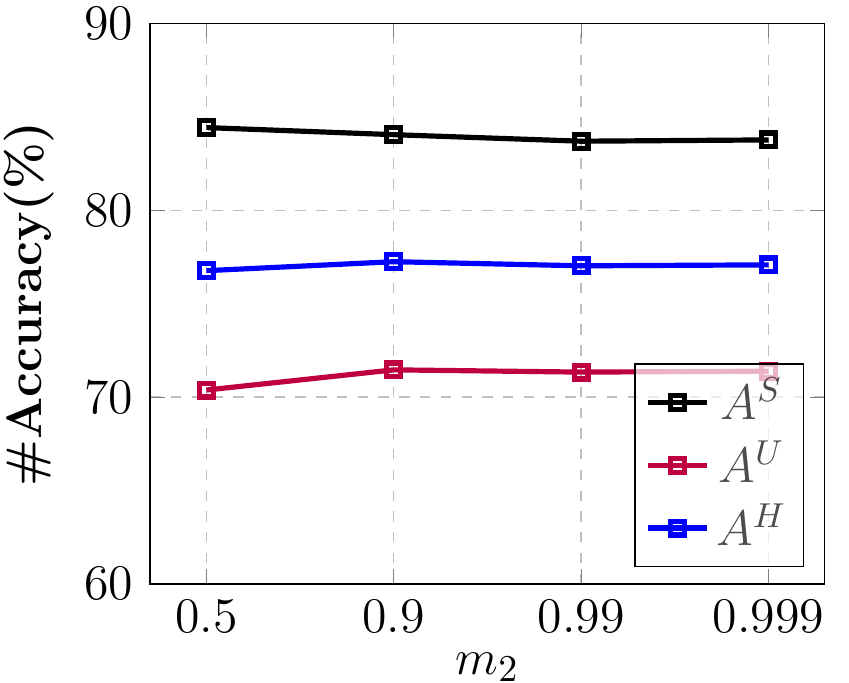}
	}
\vspace{-2ex}
	\caption{Hyperparameters \wrt~EGZSL performance on AWA2. \textbf{(a)} Effects of loss balancing coefficient $\lambda$ (Eq. \ref{eq:all}). \textbf{(b)} Effects of thresholds $\tau$ in Eq. (\ref{eq:thre_scale}). \textbf{(c), (d)} Effects of the smoothing factors $m_1$ and $m_2$ in Eq. (\ref{eq:ema}) and (\ref{eq:thre}).}
	\label{fig:para}
     \vspace{-2ex}
\end{figure}

We study the influence of the loss weighting coefficient $\lambda$, the threshold $\tau$, and the momentum coefficients $m_1$ and $m_2$, which are reported in Fig.~\ref{fig:para}. Although performance under different hyperparameter settings varies, our method is overall stable. The results are more sensitive to $\lambda$ and $m_1$ as these two parameters are related to catastrophic forgetting. In contrast, $\tau$ and $m_2$, two variables related to data selection, have a slightly smaller fluctuation in performance. We set $\lambda$ at 1, $\tau$ at 0.5, $m_1$ at 0.99, and $m_2$ at 0.9 for the best results. More experiments can be found in the supplemental.
\section{Conclusion}

In this paper, we introduce a novel and more realistic GZSL setting: Evolutionary GZSL. This setting aims to address the domain shift problem inherent in IGZSL, while maintaining greater deployability than TGZSL. EGZSL starts from the traditional learned GZSL models and gradually boosts itself by simultaneously recognizing and learning from the unlabeled test data. To evaluate the proposed EGZSL, we devise a new protocol involving random division of datasets into episodic training and testing with multiple time steps. Furthermore, we propose a method to tackle this task and present baseline results on three benchmark datasets. The results demonstrate the feasibility and superiority of our approach compared to several traditional methods.

\section*{Acknowledgement}
This work was supported in part by the National Natural Science Foundation of China under Grants 62371235 and 62072246, and in part by the Key Research and Development Plan of Jiangsu Province (Industry Foresight and Key Core Technology Project) under Grant BE2023008-2.

\bibliographystyle{named}
\bibliography{ijcai24}

\end{document}